\documentclass{article}
\usepackage[table]{xcolor}
\usepackage[dvipsnames]{xcolor}

\usepackage[preprint]{corl_2026} % Uncomment for pre-prints (e.g., arxiv); This is like ``final'', but will remove the CORL footnote.
\usepackage{booktabs}
\usepackage{graphicx}
\usepackage{hyperref}

\usepackage{xcolor}
\usepackage{amsmath}
\usepackage{amssymb}
\usepackage{multirow} 
\usepackage{array}
\usepackage{wrapfig}
\definecolor{mygray}{gray}{.93}

\title{
NativeMEM: Native Memory Compression \\
for Long-Horizon Robotic Manipulation
}

\author{
    \textbf{Ziye Wang}\textsuperscript{1}\quad
    \textbf{Modi Shi}\textsuperscript{2}\quad
    \textbf{Chaojun Ni}\textsuperscript{3}\quad
    \textbf{Jiazhi Yang}\textsuperscript{4}\quad
    \textbf{Mengdi Li}\textsuperscript{5}\quad
    \textbf{Zhizhong Su}\textsuperscript{5}\\
    \textbf{Tianwei Lin}\textsuperscript{5}\quad
    \textbf{Hongyang Li}\textsuperscript{1\dag}
    \\[4pt]
    \textsuperscript{1} The University of Hong Kong \quad
    \textsuperscript{2} Beihang University \quad
    \textsuperscript{3} Peking University\\
    \textsuperscript{4} The Chinese University of Hong Kong \quad
    \textsuperscript{5} Horizon Robotics\\
    \url{https://opendrivelab.com/NativeMEM}
    \vspace{-2.5em}
}

\begin{document}
\maketitle

\renewcommand{\thefootnote}{\fnsymbol{footnote}}
\footnotetext[2]{Corresponding author: Hongyang Li 
\href{mailto:hongyang@hku.hk}{\texttt{hongyang@hku.hk}}}

\begin{abstract}
How can pretrained Vision-Language-Action (VLA) models retain long-horizon visual histories with high-frequency updates without sacrificing efficiency? Existing approaches rely on external memory management, which restrains either the memory horizon or the reactiveness of pretrained policies. 
To this end, we present \textsc{NativeMEM}, a VLA policy that features long-term and real-time updated memory. At its core is an efficient memory encoding scheme, Native Memory Compression, which repurposes the VLA's own vision encoder to compress each historical frame from each camera view into a single token. Appended to the input sequence, these memory tokens enable the pretrained VLA to attend over long-term history with negligible latency overhead, requiring neither an external planner nor a freshly initialized memory module. To align the memory tokens with the pretrained policy, we first develop a generic memory tokenizer 
under the supervision of a frozen VLA on memory-demanding data, and then unfreeze the VLA for task-specific fine-tuning.
\textsc{NativeMEM} consistently outperforms prior methods, boosting success rates from 32.4\% to 84.0\% in simulation and up to 98.7\% on real robots, while maintaining low inference latency and GPU memory usage. Notably, \textsc{NativeMEM} exhibits high data efficiency by achieving competitive results with prior arts using only 20\% of the training data. 
\end{abstract}

% Two or three meaningful keywords should be added here

%===============================================================================

\begin{figure*}[ht]
    \centering
    \includegraphics[width=1\linewidth]{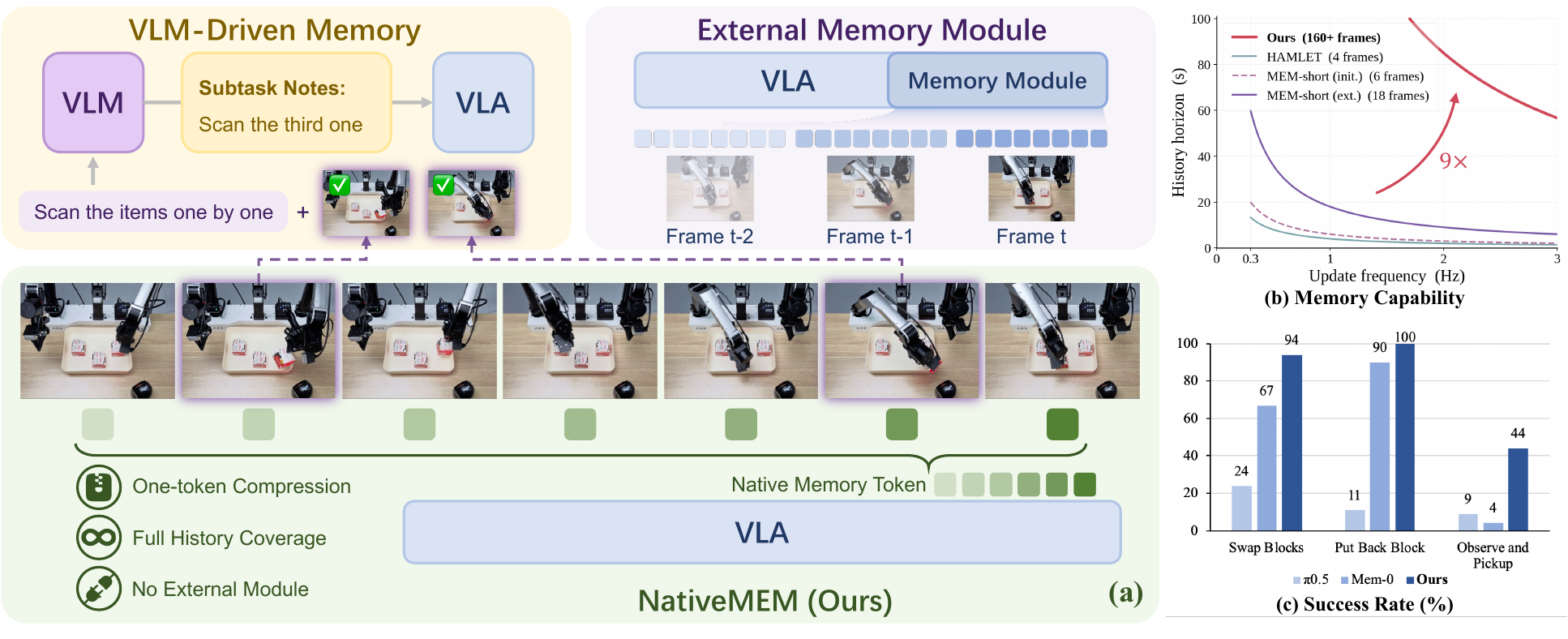} 
    \vspace{-1.2em}
    \caption{
    \textbf{\textsc{NativeMEM}} differs from prior memory-augmented VLAs that rely on VLM-generated textual notes or external memory modules. 
    (a) By repurposing the VLA's own vision encoder, it compresses each historical frame-view observation into a single native memory token, allowing the policy to condition on the full visual history through its original token sequence.
    (b) This ultra-compact representation enables minute-level histories with over 160 frames, providing a $9\times\sim40\times$ longer history horizon than prior methods.
    (c) \textsc{NativeMEM} achieves the highest success rates across memory-dependent manipulation tasks.
    }
\label{fig:teaser}
\vspace{-1em}
\end{figure*}

\section{Introduction}

Vision-Language-Action (VLA) models~\citep{brohan2023rt,openvla,pi_0,pi05,bu2025univla,team2025gigabrain,ni2026swiftvla,ye2026gigaworld,jang2025contextvla,4DVLA} extend Vision-Language Models~\citep{paligemma,smolvlm,qwen25,zhou2025vlm4d} to embodied decision-making, offering a promising path toward generalist robot control. However, most pretrained VLAs remain reactive, conditioning only on the current observation and instruction~\citep{he2026memoryarena,lauri2022partially,wu2025human}. This single-frame setup is insufficient for memory-dependent manipulation, where actions may depend on task progress, prior interactions, counts, occluded states, or failures, requiring action-relevant visual history.

To address this challenge, recent works have explored two main paradigms for memory-augmented VLAs. 
The first builds memory outside the policy, 
% typically through hierarchical VLM-VLA systems in which
where
a high-level VLM retrieves past keyframes and plans sparse subtasks for a low-level VLA controller~\citep{zhai2025memory,sridhar2025memer,torne2026mem,chen2026rmbench,wang2026hist,wang2026beyond,zeng2026helm}. While effective for extending temporal context, such systems often turn memory into text format.
% which 
%
% notes: manipulation-critical details are difficult to faithfully encode in language, and the approach may 
However, such subtask descriptions require costly annotations for training, 
and subtle details are difficult to faithfully encode in language only.
%
% The second paradigm
Another line of work builds memory 
% inside the policy
with freshly initialized modules inside the policy,
through recurrent states~\citep{li2026remem}, compressed histories~\citep{koo2025hamlet,jang2025contextvla,lin2025hif,liu2025trackvla++,nangue2025beyond,zhang2026sparse}, or retrieval-augmented memory banks~\citep{fang2025sam2act,yang20253d,gao2026gated,chen2026intent,li2026habit,li2026retrack}.
%
% Although more directly grounded in visual policy representations, these methods often introduce hand-crafted memory modules or external retrieval pipelines, 
Since these newly introduced modules are unseen during policy pretraining, this paradigm not only increases
% increasing 
the overall architectural complexity, but also introduces the risk of performance degradation of the pretrained VLA. 
More fundamentally, they face a compression dilemma: without sufficient compression, fine-grained temporal histories are too costly to retain in full; with excessive or poorly structured compression, memory may discard details critical for action. This dilemma raises a central question: \textbf{how can a VLA retain long-horizon, fine-grained histories with minimal memory cost, while remaining compatible with the pretrained policy it builds upon?}

We address this question with \textbf{Native Memory Compression}, which compresses history directly into the pretrained VLA's own visual token space. Instead of introducing a separate memory architecture, \textsc{NativeMEM} repurposes the VLA's own vision encoder to distill past observations into compact memory tokens. We push compression to its token-efficient extreme: each historical frame is represented by a single native memory token, as shown in Fig.~\ref{fig:teaser}(a). Because these tokens are produced by the same visual pathway and injected into the original token sequence used by the pretrained policy, they can be interpreted as native visual evidence rather than external memory states, allowing the VLA to access long-horizon histories through its original attention mechanisms.

To learn such memory tokens so that they are both compact and aligned with the pretrained knowledge, we introduce a two-stage training pipeline that separates learning how to summarize memory from learning how to use it for a target task. In the first stage, we freeze the pretrained VLA and train a memory tokenizer derived from its native vision encoder. Given past observations, the tokenizer compresses each frame-view pair into a single visual summary token, which is appended to the original VLA token sequence. The tokenizer is optimized through the VLA's native action prediction loss, making the summaries action-supervised rather than reconstruction-supervised. By training on a mixture of standard manipulation data and memory-demanding tasks, this stage learns a general memory tokenizer whose outputs remain aligned with the pretrained VLA's token space. In the second stage, we freeze the learned tokenizer and perform task-specific full VLA finetuning with compact memory tokens, converting a pretrained single-frame VLA into a memory-augmented policy using limited task-specific demonstrations.

We extensively evaluate \textsc{NativeMEM} on manipulation tasks demanding long-horizon and adaptive memory understanding in both the simulation and the real world. Compared with pretrained memory-free VLA policies and prior memory-augmented baselines, \textsc{NativeMEM} achieves the highest task success rates while efficiently attending to minute-level histories with compact native memory, as shown in Fig.~\ref{fig:teaser}(b) and (c). Notably, even on unseen real-robot memory tasks, \textsc{NativeMEM} converts a pretrained single-frame VLA into a memory-augmented policy using only 100 task-specific demonstrations and approximately 5 hours of finetuning. Beyond this conversion efficiency, \textsc{NativeMEM} also exhibits strong data efficiency, achieving competitive performance with prior state-of-the-art methods while using only 20\% of their training data.

\section{Related Work}

\textbf{Vision-Language-Action Models.}
Vision-Language-Action~(VLA) models~\citep{pi_0,pi05,team2025gigabrain,walloss,yang2026rise,gigaworld0,brain05,ye2025vla,li2025mimicdreamer,li2026unidrivevla,tinyvla,dreamvla} extend Vision-Language Models~(VLMs)~\citep{paligemma,smolvlm,qwen25,chen2024spatialvlm} to embodied decision making by grounding language instructions in visual observations and predicting robot actions. RT-1~\citep{brohan2022rt}, RT-2~\citep{brohan2023rt}, OpenVLA~\citep{openvla}, and $\pi_0$~\citep{pi_0} have advanced the field through large-scale robot data, vision-language pretraining, open-source generalist policies, and continuous diffusion-based action generation. However, most VLA models condition on the current observation and short-term history. Autoregressive VLAs~\citep{openvla,openvlaoft,cen2025worldvla,hu2026ar} can encode previous visual, language, or action tokens as implicit memory, but such memory remains constrained by context length and lacks an explicit mechanism for preserving task-relevant information over long-horizon interactions.

\textbf{VLM-Driven Memory with VLAs.}
To enhance memory in VLA models, recent methods often combine a high-level VLM-based memory or planning module with a low-level VLA controller~\citep{sridhar2025memer,torne2026mem,chen2026rmbench,peller2023memory}. Representative examples include MemER~\citep{sridhar2025memer}, which retrieves task-relevant keyframes to guide a VLA policy via textual instructions; MEM~\citep{torne2026mem}, which combines short-term video memory with long-term language memory for progress tracking; and Mem-0~\citep{chen2026rmbench}, which uses a VLM planner to generate subtasks for a diffusion-based executor. While effective, these methods rely on additional language-reasoning modules, increasing parameters and computation.

\textbf{Externel Memory Modules in VLAs.}
To avoid the overhead of high-level memory or planning modules, recent works introduce external memory modules 
into VLA architectures~\citep{shi2025memoryvla,li2025mapvla,neau2025grasp,guo2026chameleon}, mainly through explicit memory-bank retrieval or compressed history modeling. In the first direction, MemoryVLA~\citep{shi2025memoryvla} stores visual details and semantic tokens in a perceptual-cognitive memory bank, while MAP-VLA~\citep{li2025mapvla} builds a demonstration memory library with task-stage soft prompts. However, such methods store selective information and may omit historical details needed for subsequent action decisions.
History compression methods summarize past observations into compact representations. HAMLET~\citep{koo2025hamlet} encodes per-timestep VLM context into moment tokens, ReMem-VLA~\citep{li2026remem} uses recurrent memory queries across frames and chunks, and MEM~\citep{torne2026mem} integrates recent frames through spatial and temporal attention. Although these methods reduce the cost of full-history attention, they require architectural changes or dedicated design, limiting compatibility with pretrained VLAs.
%
% In contrast, \textsc{NativeMEM} encodes history using the pretrained VLA's vision encoder, and reuses the policy's existing attention with minimal overhead.
GridS~\citep{feng2026see} compresses visual tokens using learned anchor locations. In contrast, \textsc{NativeMEM} uses a learned memory token to compress each frame into a single token within the pretrained VLA's vision encoder and reuses the policy's existing attention with minimal overhead.

\section{Method}

\begin{figure}[t]
    \centering
    \includegraphics[width=.95\linewidth]{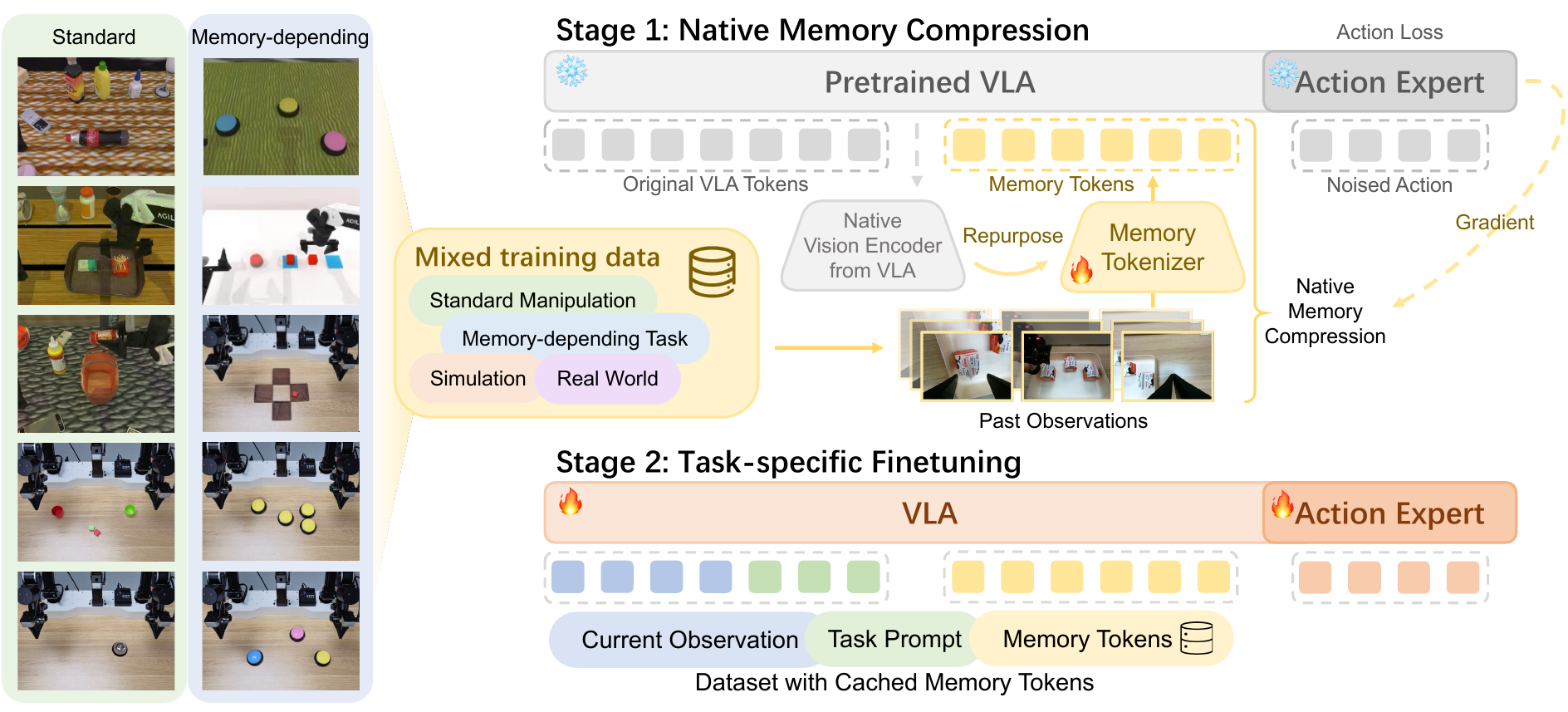}
    \caption{
    \textbf{Training pipeline of \textsc{NativeMEM}.}
    \textbf{Stage 1:} We freeze the pretrained VLA and learn a native memory tokenizer, initialized from its visual encoder, using the VLA's original action prediction loss on mixed standard and memory-dependent manipulation data. The tokenizer compresses each frame-view observation into a single action-aligned memory token.
    \textbf{Stage 2:} We cache memory tokens and finetune the VLA with the original action loss by appending compressed memory tokens to the standard current-observation and prompt tokens.
    }
    \label{fig:pipeline}
    \vspace{-0.4em}
\end{figure}

\subsection{Problem Formulation}
\label{sec:problem}

We consider robotic manipulation in a partially observable environment. At control step $t$, a pretrained single-frame VLA policy maps the current multi-view observation $\mathbf{o}_t=\{o_t^v\}_{v=1}^{V}$ (with $V$ camera views), proprioceptive state $s_t$, and language instruction $\ell$ to actions,
\begin{equation}
    \mathbf{a}_t \sim \pi_{\theta}(\cdot \mid \mathbf{o}_t, s_t, \ell),
\end{equation}
where $\theta$ denotes the parameters of the pretrained VLA.
However, such reactive policies are fundamentally limited for long-horizon manipulation, which is often non-Markovian: the correct action depends not only on the current input but on the interaction history $\mathcal{H}_t$ available before step $t$. Crucially, identical current observations can demand different actions depending on what happened before. Formally, there may exist two histories $\mathcal{H}_t \neq \mathcal{H}'_t$ such that
\begin{equation}
\begin{aligned}
    &\mathbf{o}_t = \mathbf{o}'_t,\quad
    s_t = s'_t,\quad
    \ell = \ell', \\
    &\text{but}\quad
    \mathbf{a}^{\star}_t
    (\mathcal{H}_t,\mathbf{o}_t,s_t,\ell)
    \neq
    \mathbf{a}^{\star}_t
    (\mathcal{H}'_t,\mathbf{o}_t,s_t,\ell).
\end{aligned}
\end{equation}
This is the case whenever the robot must recall previously manipulated objects, operation counts, intermediate progress, or failed attempts. In all of these, the missing ingredient is historical information, and a policy that ignores it is under-specified by construction.
We therefore convert a pretrained single-frame VLA into a memory-augmented policy that additionally conditions on this history,
\begin{equation}
    \mathbf{a}_t \sim \pi_{\theta'}(\cdot \mid \mathbf{o}_t, s_t, \ell, \mathcal{M}_t),
\end{equation}
where $\mathcal{M}_t$ is a compact memory built from the history $\mathcal{H}_t$ and $\theta'$ are the adapted parameters. The central challenge is to design $\mathcal{M}_t$ to preserve action-relevant history while staying efficient and compatible with the pretrained VLA, motivating our native memory compression scheme.

\subsection{Native Memory Compression via Action Supervision}
\label{method:stage1}

Consider a policy that maintains a visual history over $T$ seconds and updates memory at $N$ frames per second. If each historical frame from each camera view is represented by $M$ tokens, the memory length grows as
\begin{equation}
|\mathcal{M}_t| = V \cdot T \cdot N \cdot M,
\end{equation}
where $V$ is the number of camera views. This scaling reveals a fundamental \textbf{memory fidelity and efficiency trade-off}: increasing the temporal horizon $T$ or the update frequency $N$ improves the fidelity of historical context, but rapidly expands the VLA input sequence and its associated computation. Conversely, reducing the memory length improves efficiency, but risks discarding action-critical details. To retain fine-grained histories without overwhelming the VLA context, the per-frame token cost $M$ must therefore be aggressively compressed.

Our key idea, \textbf{Native Memory Compression}, is to push this compression to an extreme by setting $M=1$: each frame-view pair is summarized into a single memory token. However, compactness alone is insufficient. Since the memory tokens are consumed by a pretrained VLA, they must also be compatible with its native token space and action-generation prior. Arbitrary latent states from external modules may not be interpretable by the VLA, while summaries optimized for reconstruction or generic visual representation learning are not necessarily useful for action prediction. We therefore learn compact memory tokens directly through VLA's original action objective.

To instantiate this idea, we derive a memory tokenizer from the VLA's own visual encoder, as shown in Fig.~\ref{fig:pipeline}. Given a frame-view observation $o_\tau^v$, we initialize the memory encoder $E_{\mathrm{mem}}$ from the pretrained VLA visual encoder and introduce a learnable memory query token $q_{\mathrm{mem}}$. The query token attends to the visual patch-token sequence $\mathbf{P}_\tau^v$ and aggregates it into a single summary token:
\begin{equation}
    [\hat{q}_\tau^v, \hat{\mathbf{P}}_\tau^v]
    =
    E_{\mathrm{mem}}
    \left(
    [q_{\mathrm{mem}}, \mathbf{P}_\tau^v]
    \right),
\end{equation}
where $\hat{q}_\tau^v$ is the output memory summary. A linear memory projection maps this summary into the VLA token dimension by 
% $m_\tau^v = W_{\mathrm{mem}}\hat{q}_\tau^v$. 
\begin{equation}
    m_\tau^v = W_{\mathrm{mem}}\hat{q}_\tau^v .
\end{equation}
The resulting $m_\tau^v$ can serve as a single native visual memory token for frame $\tau$ and view $v$.

Given a set of memory frame indices $\mathcal{I}_t$, we concatenate memory tokens across time and views, preceded by a learnable memory beginning-of-sequence token $b_{\mathrm{mem}}$:
\begin{equation}
    \mathcal{M}_t =
    \left[
    b_{\mathrm{mem}},
    \{m_\tau^v \mid \tau \in \mathcal{I}_t,\; \tau \leq t,\; v=1,\ldots,V\}
    \right].
    \label{eq:memory_sequence}
\end{equation}
The memory sequence is appended to the original VLA input sequence,
\begin{equation}
    \mathbf{x}_t =
    \left[
    \mathbf{x}^{\mathrm{obs}}_t,
    \mathbf{x}^{\mathrm{prompt}}(\ell, s_t),
    \mathcal{M}_t
    \right],
    \label{eq:augmented_input}
\end{equation}
where $\mathbf{x}^{\mathrm{obs}}_t$ denotes the current observation tokens and $\mathbf{x}^{\mathrm{prompt}}(\ell, s_t)$ denotes the language and proprioceptive conditioning tokens. This introduces historical context without changing the VLA architecture: the policy attends to memory through its existing token-processing pipeline.

We learn these native memory tokens in a first-stage alignment procedure. The pretrained VLA is frozen, and only the memory branch is trainable. Given the augmented token sequence $\mathbf{x}_t$, the frozen VLA predicts the target action using its original action head, and the memory branch is optimized by the native VLA action loss:
\begin{equation}
    \min_{\phi_{\mathrm{mem}}, q_{\mathrm{mem}}, W_{\mathrm{mem}}, b_{\mathrm{mem}}}
    \mathbb{E}_{(\mathcal{H}_t,\mathbf{o}_t,s_t,\ell,\mathbf{a}_t)}
    \left[
    \mathcal{L}_{\mathrm{act}}
    \left(
    \pi_{\theta}(\cdot \mid \mathbf{o}_t, s_t, \ell, \mathcal{M}_t),
    \mathbf{a}_t
    \right)
    \right],
\end{equation}
where $\theta$ is fixed and $\phi_{\mathrm{mem}}$ denotes the parameters of the memory encoder. Since gradients can only update the memory branch, the learned tokens are encouraged to encode information that is both action-relevant and aligned with the frozen VLA's pretrained token space.

For training efficiency, we do not load the full visual history. Instead, for each training step, $\mathcal{I}_t$ consists of the first frame of the episode and a recent history window.
% ending at the current step
%
The first frame provides coarse task initialization context, while the recent window captures state changes and progress. We also include the current frame in $\mathcal{I}_t$, so that the tokenizer learns a unified frame-level summarization behavior for both current and past observations.
%
% We train the memory tokenizer on a mixture of standard manipulation data and memory-demanding tasks from both simulation and real-world robot demonstrations. Standard manipulation data preserves compatibility with the pretrained VLA's general manipulation prior, while memory-demanding tasks encourage the tokenizer to encode information that is useful beyond the current observation. The result is a general action-supervised native memory tokenizer that can be reused for downstream memory-augmented finetuning.
Our memory tokenizer is trained on a mixture of 
simulation and real-world demonstrations
covering both
standard manipulation data and memory-demanding tasks, and could be reused for downstream memory-augmented finetuning.

\subsection{Task-Specific Finetuning and Real-Time Memory Inference}
\label{method:stage2}

% After action-supervised memory token learning, we obtain a general memory tokenizer that maps each frame-view observation into a single aligned native memory token. Next, this tokenizer is used to adapt a pretrained single-frame VLA to a target memory-dependent task.

\textbf{Task-specific finetuning.}
Given a target-task demonstration dataset, we first use the learned memory tokenizer to preprocess the visual history offline. For each episode, every frame from each camera view is converted into its corresponding memory summary token $\hat{q}_\tau^v$. Since each frame-view pair is represented by only one token, this preprocessing introduces negligible storage and I/O overhead compared with storing dense visual token sequences.

For each training step $t$, we retrieve the cached summary tokens from selected frame indices $\mathcal{I}_t$ with $\tau \leq t$, and form the memory sequence following Eq.~\ref{eq:memory_sequence}. The resulting memory sequence is appended to the standard VLA input as in Eq.~\ref{eq:augmented_input}. 

We initialize the VLA backbone from the pretrained single-frame policy and load $W_{\mathrm{mem}}$ and $b_{\mathrm{mem}}$ from the first-stage memory branch. The memory tokenizer encoder is kept fixed during this stage, while the VLA backbone, action head, and memory projection are finetuned on limited target-task demonstrations using the native action prediction loss:
\begin{equation}
    \min_{\theta', W_{\mathrm{mem}}, b_{\mathrm{mem}}}
    \mathbb{E}_{(\mathbf{o}_t,s_t,\ell,\mathcal{M}_t,\mathbf{a}_t)}
    \left[
    \mathcal{L}_{\mathrm{act}}
    \left(
    \pi_{\theta'}(\cdot \mid \mathbf{o}_t, s_t, \ell, \mathcal{M}_t),
    \mathbf{a}_t
    \right)
    \right].
\end{equation}
Here, $\theta'$ denotes the finetuned VLA parameters. This preserves the standard VLA finetuning workflow while extending the policy with compact historical context.

\textbf{Real-time memory inference.}
During deployment, the memory tokenizer can operate independently from the VLA policy. As new observations arrive, the tokenizer converts them into memory tokens at a specified update frequency and maintains a compact memory queue. Since the tokenizer is derived from the original visual encoder and produces only one token per frame-view pair, memory updates can be performed with low overhead and do not require modifying the VLA inference pipeline.
When the VLA is queried for an action, the current observation, proprioceptive state, and language instruction are processed as usual. The current memory queue is simply concatenated after the original input tokens, following  Eq~\ref{eq:augmented_input}.
Thus, memory construction and policy inference are decoupled: the tokenizer can update historical memory at a high frequency, while the VLA consumes the latest compact memory sequence whenever action prediction is required.

\section{Experiments}

\subsection{Experimental Setup}

\begin{figure}[t]
    \centering
    \includegraphics[width=.99\linewidth]{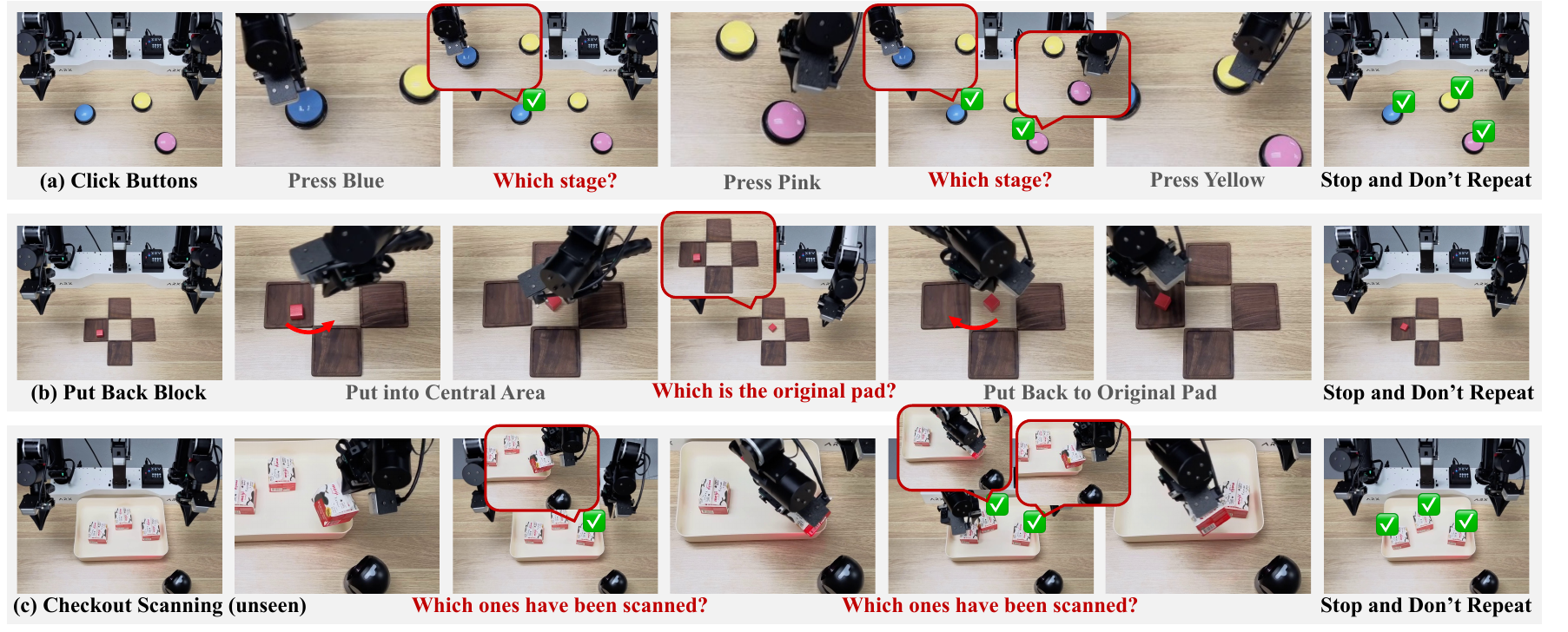}
    \caption{
    \textbf{Memory-dependent manipulation tasks.}
    In \textit{Click Buttons}, the robot must follow a specified button sequence without repeating completed presses.
    In \textit{Put Back Block}, it must remember the block's original pad after moving it to the center.
    In \textit{Grocery Checkout Scanning}, it must track which items have already been scanned and avoid duplicate or missed scans.
    }
    \label{fig:task}
    \vspace{-1.5mm}
\end{figure}

\noindent \textbf{Datasets and Tasks.}
We evaluate \textsc{NativeMEM} on memory-dependent manipulation tasks in both simulation and the real world. For real-world evaluation, we consider three tasks, as illustrated in Fig.~\ref{fig:task}. Notably, \textit{Grocery Checkout Scanning (unseen)} is not included during first-stage memory-tokenizer training. This setting evaluates whether the learned native memory representation transfers to new forms of long-horizon task progress tracking.
In simulation, we use three RMBench tasks~\citep{chen2026rmbench} and two additional button-pressing tasks. The simulated \textit{Click Buttons} task follows the same memory requirement as its real-world counterpart, while \textit{Click Buttons (Hard)} removes color distinctions between buttons, forcing the policy to rely on spatial memory and interaction history.
%
% More implementation details are included in the appendix.

\noindent \textbf{Baselines.} 
We compare \textsc{NativeMEM} with representative VLA policies that instantiate different memory-modeling paradigms. For \textit{VLM-driven memory}, we include MemER~\citep{sridhar2025memer} and Mem-0~\citep{chen2026rmbench}. For \textit{external memory modules}, we include HAMLET~\citep{koo2025hamlet} and MEM-short~\citep{torne2026mem}. 
% HAMLET compresses interaction history into moment tokens, while MEM-short conditions the policy on short-horizon visual encoded from recent video observations. 
%
For a fair comparison, we implement both HAMLET and MEM-short on the same $\pi_{0.5}$ backbone used by \textsc{NativeMEM}. Since HAMLET and MEM-short are not publicly released, we faithfully reproduce them following the official papers. MEM-short is originally designed to be retrained on large-scale data. To ensure a controlled comparison, we train it using the same demonstration trajectories available to all methods. Therefore mark them as HAMLET$^\ast$ and MEM-short$^\ast$ in Tab.~\ref{tab:results_sim} and ~\ref{tab:results_real}.

\noindent \textbf{Implementation Details.}
\textsc{NativeMEM} uses the $\pi_{0.5}$ backbone. In Stage~1, we freeze the pretrained VLA backbone and action head and train the memory branch for 50k steps with the native action prediction objective. Each sample includes the first frame and eight recent frames sampled at intervals of 8$\sim$16 steps, retaining initial state cues while limiting image I/O and memory cost. In Stage~2, we use cached memory tokens and finetune the policy from pretrained $\pi_{0.5}$ for 20k steps. All methods use the same target-task demonstrations and evaluation protocol. At inference, memory is updated every 10 control steps; the robot operates at 30~Hz with a 50-step action horizon. Each simulation and real-world task is evaluated over 100 and 50 trials, respectively.

% \noindent \textbf{Implementation Details.}
% We use $\pi_{0.5}$ as the base VLA model throughout all experiments. Following the training procedure described in Sec.~\ref{method:stage1} and ~\ref{method:stage2}, we first train the memory branch while freezing the pretrained VLA backbone and action head. The memory branch is optimized for 50,000 steps using the same action prediction objective as the base policy. After training, the memory tokenizer is applied offline to all target-task trajectories to precompute compact memory representations for each timestep and camera view. During task-specific adaptation, we initialize from the pretrained $\pi_{0.5}$ checkpoint, augment the original policy input with the cached memory tokens, and finetune the resulting \textsc{NativeMEM} for 20,000 steps. HAMLET is trained with a two-stage procedure using 50,000 steps for its first-stage moment tokens learning and 20,000 steps for task-specific finetuning, matching the training schedule of \textsc{NativeMEM}. MEM-short is trained for 50,000 steps on the same demonstrations. All methods use the same target-task demonstrations and evaluation protocol.

\begin{table*}[t]
\centering
\setlength{\tabcolsep}{6pt} % 调整列间距
\resizebox{\linewidth}{!}{%
\begin{tabular}{lcccccc}
\toprule
\textbf{Method} & \textbf{Click Buttons} &  \textbf{Click Buttons (hard)} & \textbf{Swap Blocks} & \textbf{Put Back Block} & \textbf{Observe and Pickup} & \textbf{Avg.} \\
\midrule

\rowcolor{mygray}
\multicolumn{7}{l}{\textit{Base VLA Policies}} \\
$\pi_{0.5}$~\citep{pi05} 
 & 0 & 0 & 24 & 11 & 9 & 8.8 \\
X-VLA~\citep{zheng2025x} 
 & 7 & 12 & 16 & 18 & 9 & 12.4 \\
\midrule
\rowcolor{mygray}
\multicolumn{7}{l}{\textit{VLM-Driven Memory}} \\
MemER~\citep{sridhar2025memer} 
 &12 & 8 & 18 & 12 & 2  & 10.4 \\
Mem-0~\citep{chen2026rmbench} 
& 0 & 1 & 67 & 90 & 4 & 32.4 \\
\midrule
\rowcolor{mygray}
\multicolumn{7}{l}{\textit{Externel Memory Modules}} \\
% MemoryVLA~\citep{shi2025memoryvla} & 0 & 0 & 0 & 0  & 0 & 0 \\
HAMLET$^\ast$~\citep{koo2025hamlet} & 4 & 17 & 11 & 3 & 10 & 9.0 \\
MEM-short$^\ast$~\citep{torne2026mem} & 0 & 39 & 4 & 15 & 6 & 12.8 \\
% \rowcolor{mygray}
\textbf{Ours} 
& \textbf{94} & \textbf{88} & \textbf{94} & \textbf{100} & \textbf{44} & \textbf{84.0} \\
\bottomrule
\end{tabular}}
\caption{
% Simulation 
Task success rates ($\%$) in simulation evaluation.
The benchmark includes three RMBench tasks and two additional simulated button-pressing tasks. \textsc{NativeMEM} achieves the best performance across all tasks. The best result for each task is marked in \textbf{bold}. $^\ast$ indicates reproduced baselines implemented on the same $\pi_{0.5}$ backbone.
}
\label{tab:results_sim}
\vspace{-0.4em}
\end{table*}

\begin{table*}[t]
\centering
\setlength{\tabcolsep}{6pt} % 调整列间距
\resizebox{\linewidth}{!}{%
\begin{tabular}{l*{9}{w{c}{1.2cm}}}
\toprule
\multirow{2}{*}{\textbf{Method}}
& \multicolumn{3}{c}{\textbf{Click Buttons}}
& \multicolumn{2}{c}{\textbf{Put Back Block}}
& \multicolumn{3}{c}{\textbf{Grocery Checkout Scanning (unseen)}}
& \multirow{2}{*}{\textbf{Avg.}} \\
\cmidrule(lr){2-4}
\cmidrule(lr){5-6}
\cmidrule(lr){7-9}
& \textbf{S1} & \textbf{S2} & \textbf{S3}
& \textbf{S1} & \textbf{S2}
& \textbf{S1} & \textbf{S2} & \textbf{S3}
&  \\
\midrule

$\pi_{0.5}$~\citep{pi05}
& 16 & 8 & 2
& 72 & 14
& 90 & 58 & 28
& 14.7 \\

$\pi_{0.5}$ + RTC~\citep{rtc}
& 26 & 18 & 16
& 56 & 24
& 74 & 66 & 64
& 34.7 \\

Mem-0~\citep{chen2026rmbench}
& 36 & 4 & 0
& 0 & 0
& 6 & 0 & 0
& 0.0 \\

MEM-short$^\ast$~\citep{torne2026mem}
& 72 & 52 & 40
& 6 & 0
& 72 & 56 & 52
& 30.7 \\

% \rowcolor{mygray}
\textbf{Ours}
& \textbf{100} & \textbf{98} & \textbf{96}
& \textbf{100} & \textbf{100}
& \textbf{100} & \textbf{100} & \textbf{100}
& \textbf{98.7} \\

\bottomrule
\end{tabular}}
\caption{
Real-world cumulative task success rates (\%).
S1-S3 denote cumulative success after each task stage; the last stage corresponds to the overall task success rate. \textit{Grocery Checkout Scanning} is unseen during first-stage memory-tokenizer training. The mean overall success rate is reported. The best results are marked in \textbf{bold}. $^\ast$ denotes reproduced baselines trained with the same number of task-specific demonstrations as \textsc{NativeMEM}.
}
\vspace{-1.2em}
\label{tab:results_real}
\end{table*}

\subsection{Main Results}

\noindent \textbf{Simulation Experiments.}
As shown in Tab.~\ref{tab:results_sim}, \textsc{NativeMEM} achieves the best performance across all tasks, improving the average success rate to $84.0\%$. The evaluated tasks cover two complementary memory requirements: long-range recall of early observations, as in \textit{Put Back Block} and \textit{Observe and Pickup}, and continuous progress tracking, as in \textit{Click Buttons}, \textit{Click Buttons (Hard)}, and \textit{Swap Blocks}. Existing VLM-driven or short-horizon memory methods struggle with one or both settings, while \textsc{NativeMEM} remains consistently effective, suggesting its native memory compression provides a unified representation for both recalling visual evidence and tracking task states.

We additionally fully finetune $\pi_{0.5}$ with four past frames appended at full visual-token resolution. This \textit{Native History} baseline achieves only 9/18/20\% on \textit{Swap Blocks}, \textit{Put Back Block}, and \textit{Observe and Pickup}, respectively (Tab.~\ref{tab:app_native_history}, Appendix Sec.~\ref{app:native_history}), showing that simply adding visual history does not recover \textsc{NativeMEM}'s gains.

\noindent \textbf{Real-World Experiments.}
Tab.~\ref{tab:results_real} shows that \textsc{NativeMEM} also transfers effectively to real-robot manipulation, achieving the highest overall success rate across all tasks. While RTC improves over the $\pi_{0.5}$ policy by exploiting temporal action continuity, rather than explicitly remembering past observations. Its performance still drops in later stages that require explicit historical recall. MEM-short$^\ast$ becomes less stable under limited real-world data, occasionally degrading the pretrained VLA's manipulation capability. In contrast, \textsc{NativeMEM} preserves the pretrained policy prior while adding compact history, achieving the strongest performance across all three tasks.

% On harder grocery-scanning tasks with 3/5/7 items, first-attempt success is 100/88/68\%, increasing to 100/90/74\% when retries are allowed. Several trials recover from failed scans despite no retry or recovery behavior in the 50 training demonstrations, consistent with preservation of pretrained manipulation capabilities.

\noindent \textbf{General Manipulation and Robustness.}
Tab.~\ref{tab:libero} shows that \textsc{NativeMEM} maintains comparable average performance to $\pi_{0.5}$ with both 128 and 512 memory tokens. We also evaluate \textsc{NativeMEM}'s robustness to unseen background and lighting perturbations. Without further finetuning, it achieves 67.7\% average success on the three RMBench tasks, compared with 79.3\% under clean conditions, see Appendix Sec.~\ref{app:visual_robustness} for details.

\begin{table*}[t]
\centering
\setlength{\tabcolsep}{24pt}
\resizebox{\linewidth}{!}{%
\begin{tabular}{lccccc}
\toprule
\textbf{Method}
& \textbf{Spatial}
& \textbf{Object}
& \textbf{Goal}
& \textbf{LIBERO-10}
& \textbf{Avg.} \\
\midrule
$\pi_{0.5}$
& \textbf{97.8} & \textbf{98.6} & 92.8 & 84.4 & 93.4 \\
\rowcolor{mygray}
NativeMEM$_{128}$
& 94.4 & 98.2 & 94.4 & \textbf{88.6} & 93.9 \\
\rowcolor{mygray}
NativeMEM$_{512}$
& 95.2 & 98.2 & \textbf{95.6} & 87.0 & \textbf{94.0} \\
\bottomrule
\end{tabular}}
\caption{
\textbf{General Manipulation Performance on LIBERO.}
Success rates (\%) on four LIBERO suites.
\textsc{NativeMEM} reuses the existing memory tokenizer without rerunning Stage~1 and uses the same finetuning budget as $\pi_{0.5}$ (batch size 32, 20k steps).
Subscripts denote memory-token counts. The best results are marked in \textbf{bold}.
}
\label{tab:libero}
\vspace{-0.5em}
\end{table*}

\begin{wrapfigure}{r}{0.60\linewidth}
\vspace{-1.em}
\centering
\includegraphics[width=0.95\linewidth]{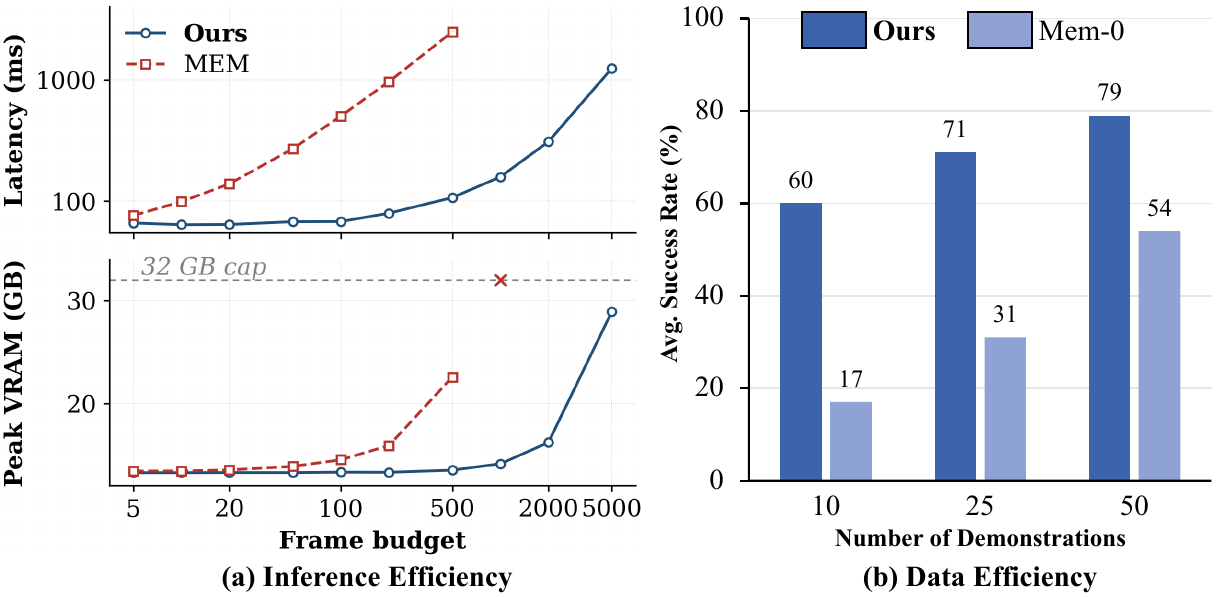}
\vspace{-0.8em}
\caption{
(a) Inference latency and memory consumption.
(b) Average data efficiency on three RMBench tasks.
}
\label{fig:efficiency}
\vspace{-0.8em}
\end{wrapfigure}

\noindent \textbf{Inference Efficiency.}
Fig.~\ref{fig:efficiency}(a) compares inference latency and peak GPU memory under increasing history length. 
%
% With one-token-per-frame native compression, 
%
Our \textsc{NativeMEM} supports up to 5,000 historical frames within a 32~GB memory budget, and still attends to about 200 frames under a 100~ms real-time latency constraint, enabling substantially longer histories while preserving the real-time reactiveness. By comparison, \textit{Native History} with 4 past frames already requires 3,840 visual tokens; 8 past frames require 6,912 tokens.

\noindent \textbf{Data-Efficiency.}
Fig.~\ref{fig:efficiency}(b) shows that \textsc{NativeMEM} adapts effectively with limited demonstrations. With only 10 demonstrations, it reaches $60\%$ average success, $3.5\times$ higher than Mem-0, and consistently outperforms Mem-0 with 25 and 50 demonstrations, reflecting the benefit of reusing pretrained VLA manipulation priors. 
%
% It only needs to learn how to condition on compact historical context, rather than relearning low-level control from limited demonstrations.

\subsection{Qualitative Analysis: What Does Native Memory Capture?}

\noindent {\raggedright\textbf{Spatial Attention of \textsc{NativeMEM}.}\par}
\begin{wrapfigure}{r}{0.59\columnwidth}
    \vspace{-2.5em}
    \centering
            \includegraphics[width=0.88\linewidth]{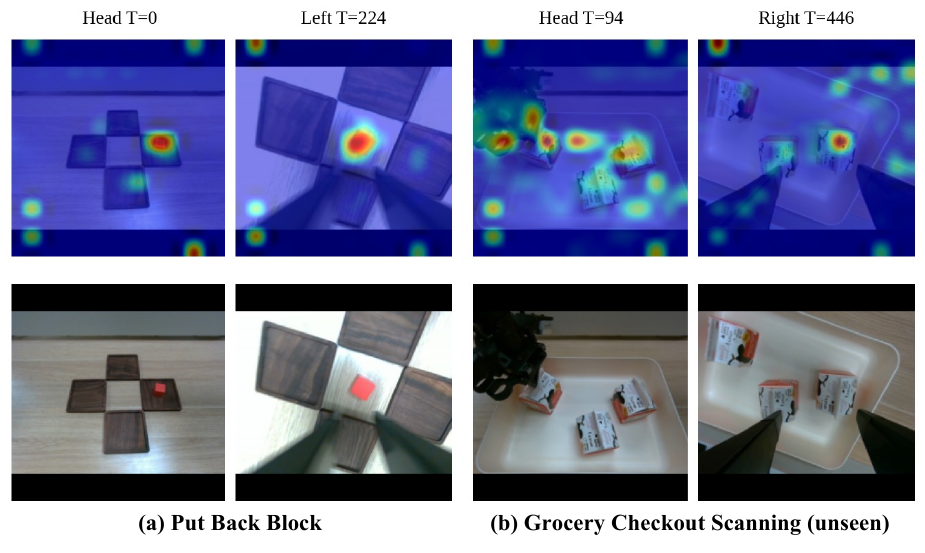}
    \vspace{-0.8em}
    \caption{\small 
    Visualization of the spatial attention used by the memory tokenizer when compressing each frame-view observation into a single memory token. The top row shows the tokenizer attention maps over input images, while the bottom row shows the corresponding raw observations.
    }
    \label{fig:spatial_attn}
    \vspace{-0.8em}
\end{wrapfigure}
\vspace{-0.4em}
To understand what is encoded into each memory token, we visualize the tokenizer's spatial attention over historical observations. As shown in Fig.~\ref{fig:spatial_attn}, the tokenizer consistently focuses on manipulation-relevant regions rather than background pixels. For \emph{Put Back Blocks}, attention concentrates on the block and its corresponding pad. Notably, this behavior also generalizes to \emph{Grocery Checkout Scanning}, which is unseen during tokenizer training: the tokenizer still attends to foreground objects, especially the item about to be grasped.

\noindent \textbf{Action Attention over the Memory Sequence.}
\begin{figure}[t]
    \centering
    \includegraphics[width=.99\linewidth]{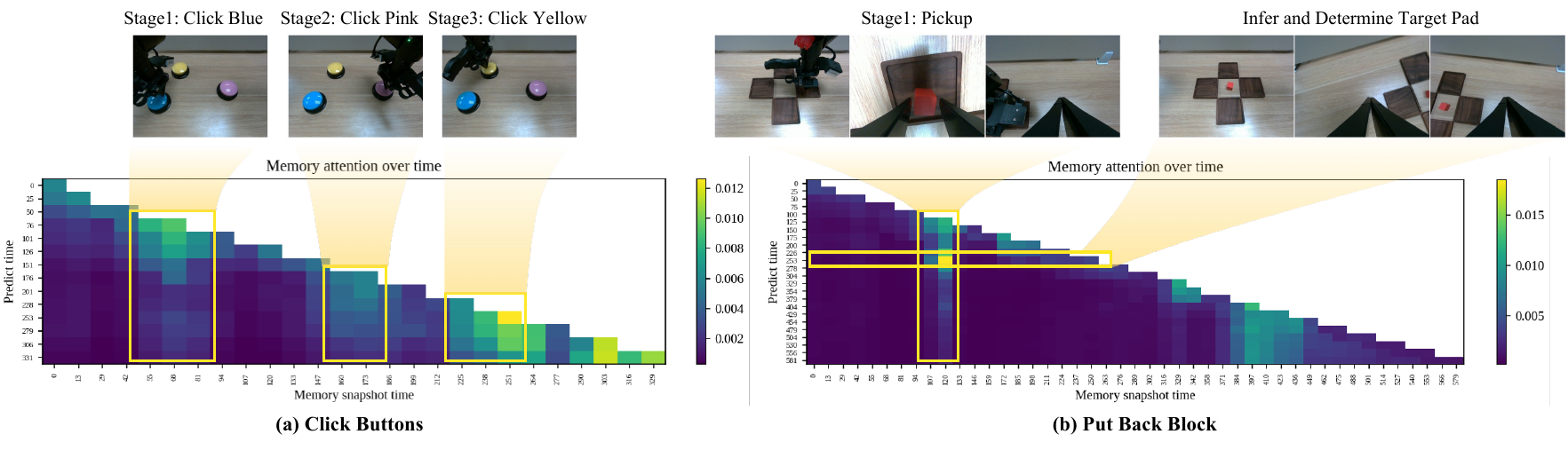}
    \vspace{-0.4em}
    \caption{
    Visualization of attention from action tokens to historical memory snapshots. The x-axis denotes the timestamp of each memory snapshot, and the y-axis denotes the action prediction timestep. Each row shows which memories are used for a given action prediction time, while each column indicates how a specific memory is attended by future predictions. The attention in \emph{Click Buttons} follows the three clicking stages, while in \emph{Put Back Block} it concentrates on the pickup moment needed to infer the target pad.
    }
    \label{fig:temporal_attn}
    \vspace{-0.4em}
\end{figure}
To examine how \textsc{NativeMEM} uses history during action generation, we visualize the attention from action tokens to memory tokens across inference time. Fig.~\ref{fig:temporal_attn} shows that the policy attends to task-relevant moments rather than simply the most recent observations. In \emph{Click Buttons}, high-attention memories align with the three timesteps when individual buttons were pressed. In \emph{Put Back Block}, attention concentrates on the moment when the block was lifted from its original pad, which is critical for deciding where to return it.

\subsection{Ablation Study}

\begin{table*}[t]
\centering
\setlength{\tabcolsep}{6pt} % 调整列间距
\resizebox{\linewidth}{!}{%
\begin{tabular}{lccccccc}
\toprule
\textbf{Method} & \textbf{Click Button} &  \textbf{Click Button (hard)} & \textbf{Swap Blocks} & \textbf{Put Back Block} & \textbf{Observe and Pickup} & \textbf{Avg.}\\
\midrule
Unfrozen VLA & 38 & 24 & 45 & 0 & 9 & 23.2 \\
w/o Stage1 & 94 & 80 & 92 & 17 & 7 & 58.0 \\
Sparse update (0.5Hz) & 83 & 63 & 91 & 24 & 9 & 53.8 \\
Short horizon (2s) & 26 & 87 & 0 & 18 & 39 & 34.0 \\
Short horizon (4s) & 69 & 87 & 0 & 18 & 37 & 43.0 \\
Short horizon (6s) & 80 & 88 & 0 & 18 & 44 & 46.0 \\
\rowcolor{mygray}
\textbf{Ours} 
& \textbf{94} & \textbf{88} & \textbf{94} & \textbf{100} & \textbf{44} & \textbf{84.0} \\
\bottomrule
\end{tabular}}
\caption{
\textbf{Ablation Study on Native Memory Compression and its Temporal Coverage.} 
The first two variants evaluate the memory-compression learning, while the remaining variants examine the importance of long-horizon and fine-grained history. The best results are marked in \textbf{bold}.
}
\label{tab:ablation}
\vspace{-0.8em}
\end{table*}

\noindent \textbf{Native Memory Compression.}
As shown in Tab.~\ref{tab:ablation}~(lines 1 and 2), when the VLA is unfrozen during memory-compression learning, performance drops to $23.2\%$. The model reduces the action loss by directly adapting the pretrained VLA itself rather than forcing the memory branch. After skipping Native Memory Compression, mean-pooled vision-encoder features are used as memory tokens. The results show that such generic visual features can capture coarse historical context, but fail to preserve details required by tasks such as \textit{Put Back Block} and \textit{Observe and Pickup}. 
These results indicate that Native Memory Compression is essential for distilling task-relevant historical cues, enabling the policy to recover critical information needed for downstream action prediction.

\noindent \textbf{Temporal Memory Coverage.} 
Sparse update at $0.5$Hz broadly reduces success rates, showing that the policy needs fine-grained temporal evidence to track interaction states. Short horizons degrade tasks whose critical information falls outside the retained window, such as \textit{Swap Blocks} and \textit{Put Back Block}.
Results in Tab.~\ref{tab:ablation}~(lines 4$\sim$6) show that \textsc{NativeMEM}'s performance relies on both dense temporal coverage and long-horizon retention: it updates memory frequently to capture fine-grained interaction changes, while its compact tokens make minute-level history retention scalable.

\section{Conclusion}

% Our contributions are threefold:
% \textbf{(1)}
% We propose \textsc{NativeMEM}, a memory-augmented VLA that uses Native Memory Compression to turn each historical frame from each camera view into a single visual memory token, enabling minute-level fine-grained visual histories.
% \textbf{(2)} 
% We introduce a two-stage training pipeline that first learns a memory tokenizer via the pretrained VLA's native action objective, aligning compact memory tokens with visual-action priors and enabling data-efficient task-specific finetuning from limited demonstrations.
% \textbf{(3)} 
%   \textsc{NativeMEM} improves average success rates from 32.4\% to \textbf{84.0\%} in simulation and from 34.7\% to \textbf{98.7\%} on real robots, and matches leading memory-designed methods with only \textbf{20\%} of the training data.

We present \textsc{NativeMEM}, which enables pretrained single-frame VLAs to retain long-horizon, fine-grained visual histories through \textbf{Native Memory Compression}. By repurposing the VLA's own vision encoder, \textsc{NativeMEM} achieves one-token-per-frame compression. We further introduced a two-stage training pipeline that first learns an action-supervised memory tokenizer aligned with the pretrained VLA's visual-action priors, and then performs task-specific finetuning with limited demonstrations. Across simulation and real-world manipulation tasks, \textsc{NativeMEM} substantially improves average success rates from 32.4\% to 84.0\% in simulation and from 34.7\% to 98.7\% on real robots, while matching leading memory-designed methods with only 20\% of the training data.

\clearpage
% The acknowledgments are automatically included only in the final and preprint versions of the paper.
\acknowledgments{
This work was supported by the Jockey Club AIS Lab, the IDS Seed Fund, and research sponsorship from Horizon Robotics. We also thank Hai Zhang and Bangjun Wang for their valuable comments.
}

%===============================================================================

% no \bibliographystyle is required, since the corl style is automatically used.
\bibliography{example}  % .bib

\clearpage
\appendix
\begin{center}
    {\LARGE\bfseries Appendix\par}
\end{center}
% \vspace{1em}

\begin{figure}[ht]
    \centering
    \includegraphics[width=.95\linewidth]{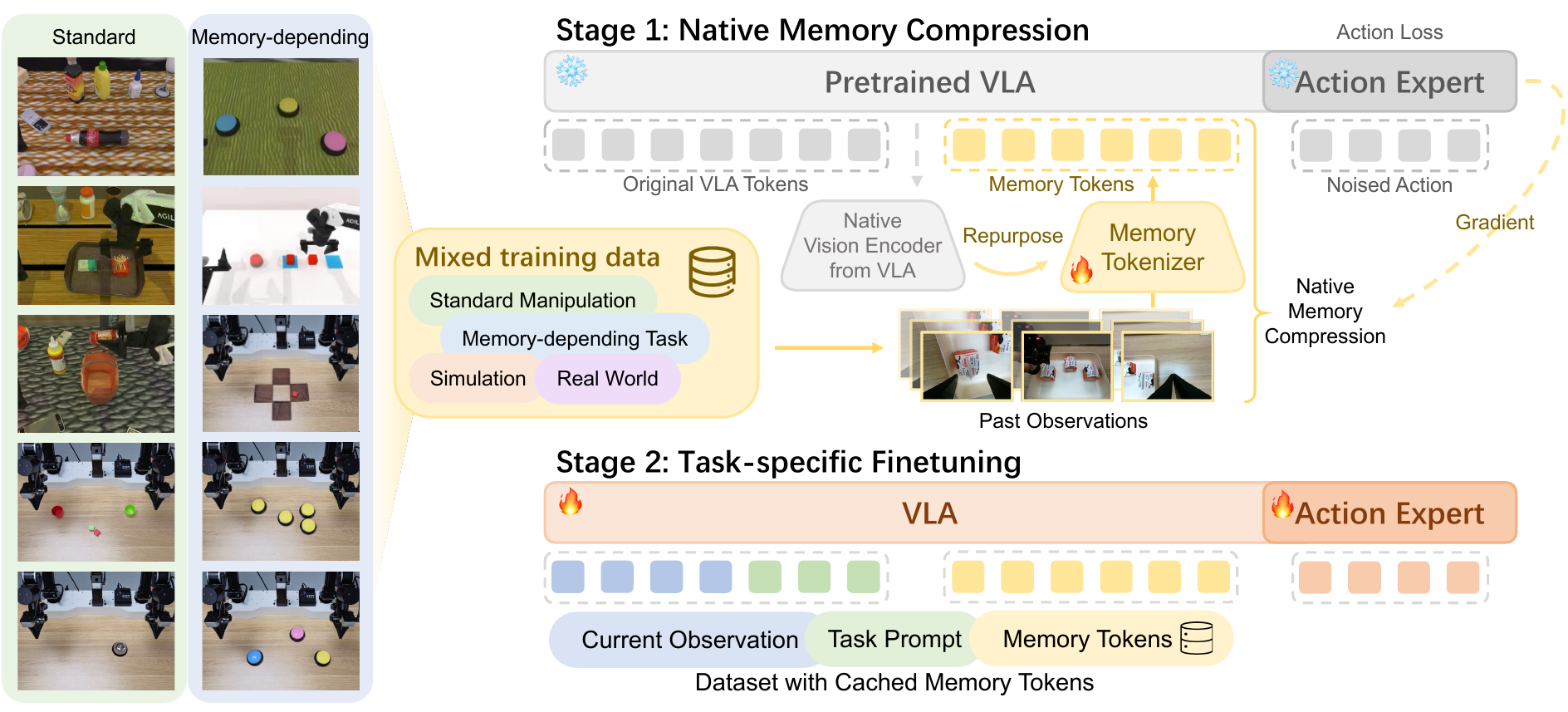}
    \caption{
    \textbf{Training pipeline of \textsc{NativeMEM}.}
    \textbf{Stage 1:} We freeze the pretrained VLA and learn a native memory tokenizer, initialized from its visual encoder, using the VLA's original action prediction loss on mixed standard and memory-dependent manipulation data. The tokenizer compresses each frame-view observation into a single action-aligned memory token.
    \textbf{Stage 2:} We cache memory tokens and finetune the VLA with the original action loss by appending compressed memory tokens to the standard current-observation and prompt tokens.
    }
    \label{fig:pipeline}
    \vspace{-0.8em}
\end{figure}

\section{Implementation Details}

\noindent \textbf{Pipeline.} Fig.~\ref{fig:pipeline} illustrates the two-stage training pipeline of \textsc{NativeMEM}. The first stage learns a native memory tokenizer while keeping the pretrained VLA frozen. The second stage uses the learned tokenizer to cache compact memory summaries for target-task demonstrations and finetunes the VLA to condition on these historical memory tokens for action prediction. 

\noindent \textbf{Stage 1: Training the native memory tokenizer.}
In the first stage, we train the native memory tokenizer to compress each frame-view observation into a compact memory token. The tokenizer is initialized from the pretrained VLA visual encoder to keep its outputs aligned with the policy's visual representation space. During training, the pretrained VLA is frozen and only the memory branch is updated using the VLA action prediction loss, without reconstruction objectives or additional memory supervision. This action-supervised objective encourages the tokenizer to preserve information useful for downstream control. We train on approximately 4,000 trajectories from both simulation and real-world environments, mixing standard manipulation demonstrations with memory-dependent ones. The former maintains compatibility with the pretrained VLA's manipulation prior, while the latter encourages encoding history unavailable from the current observation alone. In our implementation, this stage is trained for 50,000 steps with a batch size of 16 and a learning rate of $5\times10^{-5}$, taking approximately 16 hours on 4 NVIDIA H100 GPUs.

% \noindent \textbf{Stage 1: Training the native memory tokenizer.} In the first stage, we train the native memory tokenizer to compress each frame-view observation into a single compact memory token. The tokenizer is initialized from the visual encoder of the pretrained VLA, so that its output remains close to the visual representation space already used by the policy. During this stage, all parameters of the pretrained VLA are frozen, and only the memory branch is updated. We optimize this branch using the action prediction loss of the VLA, without introducing any reconstruction objective or additional memory-specific supervision. This action-supervised objective encourages the tokenizer to preserve information that is directly useful for downstream robot action prediction. We train the tokenizer on a mixture of standard manipulation demonstrations and memory-dependent demonstrations from both simulation and real-world data. Standard manipulation data helps maintain compatibility with the pretrained VLA's general manipulation prior, while memory-dependent data encourages the tokenizer to encode historical information that cannot be inferred from the current observation alone. In our implementation, this stage is trained for 50,000 steps with a batch size of 16 and a learning rate of $5\times10^{-5}$. Training takes approximately 16 hours on 4 NVIDIA H100 GPUs. The training data for this stage consists of 2,000 episodes collected from both simulation and real-world environments. 

\noindent \textbf{Stage 2: Finetuning the VLA with cached memory summaries.} In the second stage, we use the trained memory tokenizer from Stage 1 to preprocess target-task demonstrations. For each trajectory, every frame from each camera view is converted into a single compact memory summary and cached offline. Since each frame-view pair contributes only one token-level feature, this preprocessing introduces negligible storage and I/O overhead compared with storing raw images or dense visual token sequences. During finetuning, we retrieve cached memory summaries according to the predefined temporal sampling interval and append the corresponding memory tokens to the standard VLA input sequence. The VLA is then finetuned with the same original action prediction loss, allowing the policy to learn how to attend to compact historical context when predicting actions. Unlike Stage 1, where the pretrained VLA is frozen, Stage 2 updates the VLA parameters for the target task. The memory tokenizer encoder is kept fixed because its outputs are cached offline, while the VLA parameters and the lightweight memory projection are adapted during finetuning. This design avoids repeatedly running the tokenizer during VLA training and preserves the standard VLA finetuning workflow. We finetune the policy for 20,000 steps with a batch size of 32. Specifically, we use 50 episodes for simulation-based finetuning and 100 episodes for real-world finetuning. This stage takes approximately 5 hours on 2 NVIDIA H100 GPUs.

\section{Baselines}

We compare \textsc{NativeMEM} with representative VLA policies that instantiate different memory-modeling paradigms, including VLM-driven memory methods and methods with external memory modules. For a fair comparison, all reproduced baselines are implemented on the same $\pi_{0.5}$ backbone used by \textsc{NativeMEM} whenever applicable.

\noindent \textbf{MemER.}
MemER~\cite{sridhar2025memer} is a VLM-driven memory method that uses a vision-language model to summarize and retrieve task-relevant historical information. It maintains memory at the semantic level and provides additional contextual information to the policy for decision-making. 

\noindent \textbf{Mem-0.}
Mem-0~\cite{chen2026rmbench}  is a hierarchical planning and execution framework. The planning module generates high-level subtasks from instructions, observations, and memory. The execution module produces low-level actions using current observations and retrieved memory. A subtask classifier determines completion and stores key frames to update memory. 

\noindent \textbf{HAMLET$^\ast$.}
HAMLET$^\ast$~\cite{koo2025hamlet} augments VLA models with history awareness by introducing moment tokens that encode timestep-wise perceptual information. These tokens are refined via time-contrastive learning and aggregated through a memory module to form a historical context. The resulting memory is used to improve action prediction, especially for long-horizon tasks. 
% HAMLET$^\ast$ is trained with a two-stage procedure using 50,000 steps for its first-stage moment tokens learning and 20,000 steps for task-specific finetuning, matching the training schedule of \textsc{NativeMEM}

\noindent \textbf{MEM-short$^\ast$.}
MEM-short$^\ast$~\cite{torne2026mem} introduces a short-horizon visual memory for vision-language-action models by using a video encoder that compresses multiple past frames into a compact representation. It extends Vision Transformers with temporal attention while keeping spatial encoding intact, allowing efficient aggregation of recent observations. This design reduces inference cost while preserving fine-grained motion and dynamic information needed for short-term reasoning. 
% MEM-short$^\ast$ is trained for 50,000 steps on the same demonstrations.

\section{Additional Experimental Results}
\label{app:additional_results}

We provide additional evaluations of \textsc{NativeMEM} under
unseen visual perturbations, with uncompressed visual history,
and on more challenging memory-dependent tasks. We also examine
its behavior under dynamic human intervention and failure recovery.
These experiments complement the main results by examining
robustness, the benefits of native memory compression, and the
capabilities and limitations inherited from the pretrained VLA.
Simulation and real-world evaluations use 100 and 50 trials,
respectively.

\subsection{Robustness to Unseen Visual Perturbations}
\label{app:visual_robustness}

We evaluate \textsc{NativeMEM} on three RMBench tasks under unseen
background and lighting perturbations, without further finetuning.
As shown in Table~\ref{tab:app_robustness}, average success decreases
from 79.3\% in clean scenes to 67.7\% under the harder conditions.
Although performance decreases across all three tasks,
\textsc{NativeMEM} retains substantial task performance under
these visual changes.

\begin{table*}[t]
\centering
\setlength{\tabcolsep}{20pt}
\resizebox{\linewidth}{!}{%
\begin{tabular}{lcccc}
\toprule
\textbf{Setting}
& \textbf{Swap Blocks}
& \textbf{Put Back Block}
& \textbf{Observe and Pickup}
& \textbf{Avg.} \\
\midrule
Clean & 94 & 100 & 44 & 79.3 \\
Hard  & 85 & 86  & 32 & 67.7 \\
\bottomrule
\end{tabular}}
\caption{
\textbf{Robustness to Unseen Visual Perturbations.}
Success rates (\%) under clean conditions and unseen background
and lighting perturbations (Hard), without additional finetuning.
}
\label{tab:app_robustness}
\vspace{-0.8em}
\end{table*}

\subsection{Comparison with Uncompressed Visual History}
\label{app:native_history}

To examine whether the improvements arise simply from finetuning
the base VLA with historical observations, we fully finetune
$\pi_{0.5}$ while directly appending $K$ past frames at full
visual-token resolution. With $K=4$, this \textit{Native History}
baseline requires $5\times3\times256=3{,}840$ visual tokens,
accounting for the current frame and four historical frames from
three camera views. Despite this token cost, it performs
substantially below \textsc{NativeMEM} on all three tasks
(Table~\ref{tab:app_native_history}). Increasing the history to $K=8$ requires
$9\times3\times256=6{,}912$ visual tokens and exceeds the training
capacity. In contrast,
\textsc{NativeMEM} efficiently conditions on more than 160
historical frames. These results show that directly adding a
short, uncompressed visual history does not reproduce the gains
of native memory compression.

\begin{table*}[t]
\centering
\setlength{\tabcolsep}{20pt}
\resizebox{\linewidth}{!}{%
\begin{tabular}{lccc}
\toprule
\textbf{Method}
& \textbf{Swap Blocks}
& \textbf{Put Back Block}
& \textbf{Observe and Pickup} \\
\midrule
$\pi_{0.5}$             & 24 & 11  & 9 \\
Native History ($K=4$) & 9  & 18  & 20 \\
\textsc{NativeMEM}      & \textbf{94}
                       & \textbf{100}
                       & \textbf{44} \\
\bottomrule
\end{tabular}}
\caption{
\textbf{Comparison with Uncompressed Visual History.}
Success rates (\%) for the base policy, a fully finetuned policy
with four full-resolution historical frames, and
\textsc{NativeMEM}. The best results are marked in \textbf{bold}.
}
\label{tab:app_native_history}
\vspace{-0.8em}
\end{table*}

\subsection{Harder Grocery Scanning Tasks}
\label{app:hard_scanning}

We further evaluate \textsc{NativeMEM} on grocery-scanning tasks
with three, five, and seven items. These tasks are unseen during
Stage~1, and task-specific finetuning uses only 50 demonstrations.
Table~\ref{tab:app_scanning} reports first-attempt success
(\textit{Once}) and success when retries are allowed
(\textit{Retry}). As the number of items increases, first-attempt
success decreases from 100\% to 88\% and 68\%, reflecting the
greater demands on manipulation and progress tracking. Allowing
retries raises success to 90\% for five items and 74\% for seven
items. No retry or failure-recovery behavior appears in the 50 training
demonstrations.

\begin{table*}[t]
\centering
\setlength{\tabcolsep}{18pt}
\resizebox{\linewidth}{!}{%
\begin{tabular}{lcccccc}
\toprule
\multirow{2}{*}{\textbf{Method}}
& \multicolumn{2}{c}{\textbf{Scan 3 Items}}
& \multicolumn{2}{c}{\textbf{Scan 5 Items}}
& \multicolumn{2}{c}{\textbf{Scan 7 Items}} \\
\cmidrule(lr){2-3}
\cmidrule(lr){4-5}
\cmidrule(lr){6-7}
& \textbf{Once} & \textbf{Retry}
& \textbf{Once} & \textbf{Retry}
& \textbf{Once} & \textbf{Retry} \\
\midrule
\textsc{NativeMEM}
& 100 & 100
& 88  & 90
& 68  & 74 \\
\bottomrule
\end{tabular}}
\caption{
\textbf{Grocery Scanning with Increasing Task Difficulty.}
Success rates (\%) on tasks involving three, five, and seven
items. Once denotes first-attempt success, and Retry denotes
success when retries are allowed.
}
\label{tab:app_scanning}
\vspace{-0.8em}
\end{table*}

\begin{figure}[t]
    \centering
    \includegraphics[width=.95\linewidth]{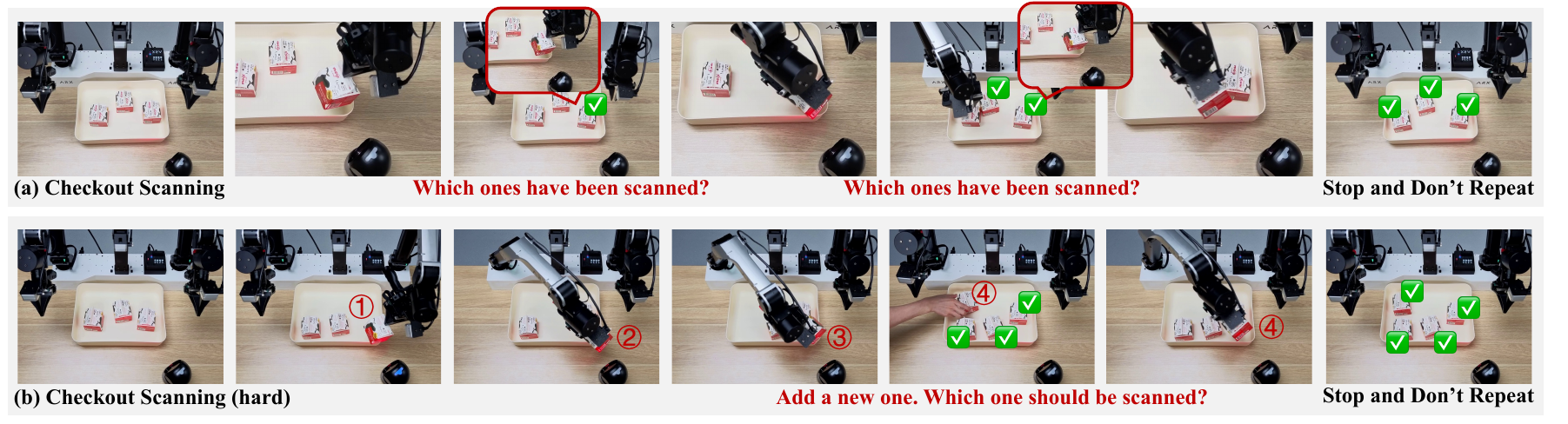}
    \caption{
    \textbf{Checkout scanning under dynamic human intervention.}
    \textbf{Top:} In the standard setting, the robot scans all
    objects and stops without repeating previously scanned items.
    \textbf{Bottom:} In the hard setting, after the initial objects
    have been scanned, a human inserts a new object at an arbitrary
    position on the left side of the workspace. The robot must
    identify the newly introduced object, scan only the unscanned
    object, and then stop without repetition.
    }
    \label{fig:hard_task}
    \vspace{-0.8em}
\end{figure}

\subsection{Human-Intervention Test}
\label{app:human_intervention}

To further evaluate whether \textsc{NativeMEM} can maintain and
utilize task-relevant history under dynamic scene changes, we
design a human-intervention stress test, as shown in
Fig.~\ref{fig:hard_task}. The robot first scans the original
objects and must remember which items have been processed.
Afterward, a human inserts an additional object at an arbitrary
position on the left side of the workspace. The policy must
distinguish the newly introduced object from previously scanned
ones and scan only the unprocessed item. This setting is
challenging because the current observation alone does not reveal
which objects have already been scanned, especially when their
spatial arrangements change.

As shown in Fig.~\ref{fig:hard_task} and the supplementary video
\texttt{NativeMEM.mp4}, \textsc{NativeMEM} identifies the newly
inserted object as unscanned, scans it, and stops without
revisiting previously processed items. The video also shows a
failure-recovery case: the robot initially fails to grasp the
third item, then retries and completes the scan. Consequently,
four attempts occur while only three items are successfully
scanned. A policy that simply counts attempts would conflate
failed attempts with successfully processed items. In contrast,
when the fourth item is introduced, \textsc{NativeMEM} identifies
it as unscanned and performs the additional scan.

This behavior suggests that the memory representation supports
outcome-aware task progress tracking, rather than merely counting
action attempts. Moreover, recovery behavior is absent from the
target-task finetuning demonstrations. These observations are
consistent with \textsc{NativeMEM} preserving the pretrained VLA's
general manipulation prior while extending it with long-horizon
memory. This test provides qualitative behavioral evidence,
complementing the quantitative scanning results in
Table~\ref{tab:app_scanning}.

\subsection{Additional RMBench Tasks and Failure Cases}
\label{app:additional_rmbench}

Table~\ref{tab:app_rmbench} reports results on the remaining
RMBench tasks. \textsc{NativeMEM} achieves 98\% success on
\textit{Rearrange Block} and 69\% on \textit{Swap T}, compared
with 89\% and 14\% for Mem-0. However, performance remains low on
\textit{Press Button}, which requires reading written numbers
and pressing buttons the specified number of times.

This failure case illustrates a limitation of memory augmentation:
retaining task-relevant history does not by itself supply
capabilities absent from the underlying VLA. With external number
recognition, \textsc{NativeMEM} reaches 5\% success on
\textit{Press Button}, while all evaluated VLM+VLA baselines
remain at 0\% under the same setting. In a simplified setting
that requires pressing only one button, \textsc{NativeMEM}
achieves 32\% success. These results suggest that both recognition
and reliable execution of the requested press count remain
challenges.

\begin{table*}[t]
\centering
\setlength{\tabcolsep}{14pt}
\resizebox{\linewidth}{!}{%
\begin{tabular}{lcccc}
\toprule
\textbf{Method}
& \textbf{Rearrange Block}
& \textbf{Swap T}
& \textbf{Press Button}
& \textbf{Press Button (easy)} \\
\midrule
Mem-0
& 89 & 14 & 0 & -- \\
\textsc{NativeMEM}
& \textbf{98} & \textbf{69} & \textbf{5} & 32 \\
\bottomrule
\end{tabular}}
\caption{
\textbf{Additional RMBench Tasks and Failure Cases.}
Success rates (\%) on the remaining RMBench tasks. Press Button
uses external number recognition; the easy setting requires
pressing only one button. A dash indicates an unreported result.
The best available results for tasks with both methods evaluated
are marked in \textbf{bold}.
}
\label{tab:app_rmbench}
\vspace{-0.8em}
\end{table*}

\section{Supplementary Video}

We provide a supplementary video, \texttt{NativeMEM.mp4}, to illustrate the behavior and analysis of \textsc{NativeMEM}. The video contains the following segments:

\begin{itemize}
\item \textbf{0:00--0:35: Checkout scanning with real-time dashboards.}
\textsc{NativeMEM} first scans three items in sequence. When an additional item is manually inserted into the scene, the policy recognizes it as unscanned, performs the corresponding scan, and stops without repeating previously scanned items. The segment also shows a real-time dashboard reporting the accumulated historical frames included in memory and the inference latency, illustrating that \textsc{NativeMEM} maintains long-horizon visual history with low runtime overhead.

\item \textbf{0:35--0:56: Method overview.}
We visualize the overall architecture of \textsc{NativeMEM}, including native memory compression, memory-token insertion, and memory-augmented VLA finetuning.

\item \textbf{0:56--1:47: Memory-dependent manipulation.}
We show representative examples from Click Buttons, Put Back Block, and Checkout Scanning, together with memory-tokenizer attention heatmaps that highlight the visual evidence encoded into memory.

\item \textbf{1:47--2:34: Comparison with baselines.}
We compare \textsc{NativeMEM} with multiple baseline methods under the same task settings, showing that compact native memory enables more reliable long-horizon manipulation.

\item \textbf{2:34--3:00: Failure recovery.}
We present a recovery case in Checkout Scanning where the robot fails to grasp an item on the first attempt, retries successfully, and still maintains the correct task progress. This demonstrates that \textsc{NativeMEM} tracks successful task outcomes rather than merely counting action attempts.

\end{itemize}

\end{document}